\documentclass{article} 
\usepackage{nips15submit_e,times}
\usepackage{hyperref}
\usepackage{url}
\usepackage{amsfonts}

\usepackage{bm}
\usepackage{algorithm}
\usepackage[noend]{algorithmic}
\usepackage{amsmath}
\usepackage{mathtools}
\newtheorem{mydef}{Definition}
\usepackage{graphicx}
\usepackage{subfigure}
\usepackage{tabularx}
\usepackage{wrapfig}
\usepackage{float}
\usepackage[backend=bibtex,firstinits=true,maxbibnames=99,sorting=none]{biblatex}

\usepackage{color}

\newcommand{\chrisx}[1]{{\color{red}{\bf\sf [CJF: #1]}}}

\title{Streaming Gibbs Sampling for LDA Model}

\author{
Yang Gao \\
Electrical Engineering and Computer Science\\
University of California Berkeley\\
Berkeley, CA 94710 \\
\texttt{yg@eecs.berkeley.edu} \\
\And
Jianfei Chen, Jun Zhu \\
Dept. of Comp. Sci \& Tech; TNList Lab, \\
State Key Lab of Intell. Tech \& Sys. \\
Tsinghua University, Beijing, 100084, China \\
\texttt{chenjf10@mails.tsinghua.edu.cn} \\
\texttt{dcszj@mail.tsinghua.edu.cn} \\ 
}

\nipsfinalcopy 

\bibliography{nips2015_SGS}

\begin{document}

\maketitle

\begin{abstract}
Streaming variational Bayes (SVB) is successful in learning LDA models in an online manner. However previous attempts toward developing online Monte-Carlo methods for LDA have little success, often by having much worse perplexity than their batch counterparts. We present a streaming Gibbs sampling (SGS) method, an online extension of the collapsed Gibbs sampling (CGS). Our empirical study shows that SGS can reach similar perplexity as CGS, much better than SVB. Our distributed version of SGS, DSGS, is much more scalable than SVB mainly because the updates' communication complexity is small.
\end{abstract}

\section{Introduction}
\label{intro}
Topic models such as Latent Dirichlet Allocation (LDA) ~\cite{blei2003latent} have gained increasing attention. LDA provides interpretable low dimensional representation of documents, uncovering the latent topics of the corpus. The model has been proven to be useful in many fields, such as natural language processing, information retrieval and recommendation systems~\cite{mitchell2008vector, naveed2011bad}. Companies such as Google~\cite{liu2011plda+}, Yahoo!~\cite{ahmed2012scalable}, and Tencent~\cite{wang2014peacock} have taken advantage of the model extensively. LDA has gradually become a standard tool to analyze documents in a semantic perspective.

With the Internet generating a huge amount of data each day, accommodating new data requires periodic retraining using traditional batch inference algorithms such as variational Bayes (VB) ~\cite{blei2003latent}, collapsed Gibbs sampling (CGS) ~\cite{griffiths2004finding} or their variants ~\cite{hoffman2013stochastic, foulds2013stochastic, patterson2013stochastic}, which might be a waste of both computational and storage resources, due to the need of recomputing and storing all historical data. It is better to learn the model \emph{incrementally}, with online (streaming) algorithms. An online algorithm is defined as the method that learns the model in a single pass of the data and can analyze a test document at any time during learning. 

Stochastic variational inference (SVI) is an offline stochastic learning algorithm that has enjoyed great experimental success on LDA~\cite{broderick2013streaming}. Although online methods often perform worse than offline ones, streaming variational Bayes (SVB) \cite{broderick2013streaming} achieves the same performance as the offline SVI, which is impressive. However, these variational methods need to make unwarranted mean-field assumptions and require model-specific derivations. In contrast, Monte Carlo methods (e.g., CGS) are generally applicable and asymptotically converge to the target posterior. By exploring the sparsity structure, CGS has been adopted in many scalable algorithms for LDA~\cite{yuan2014lightlda,ahmed2012scalable}.
However, the attempts towards developing streaming Monte Carlo methods for LDA have little success, often achieving worse performance than CGS. 
For instance, the perplexity of OLDA~\cite{alsumait2008line} on NIPS dataset is much worse than CGS; and the particle filter approach \cite{canini2009online} could  only achieve 50\% performance of CGS, in terms of the normalized mutual information on labeled corpus. 



In this paper, we fill up this gap by presenting a streaming Gibbs sampling (SGS) algorithm. SGS naturally extends the collapsed Gibbs sampling to the online learning setting. We empirically verify that, with weight decay, SGS can reach similar inference quality as CGS. 
We further present a distributed version of SGS (DSGS) in order to deal with large-scale datasets on multiple compute nodes. Empirical results demonstrate that the distributed SGS achieves comparable inference quality as the non-distributed one, while with dramatic scaling-up. 
Since DSGS can be implemented with sparse data structures, demanding much lower communication bandwidth, it is more scalable than SVB. Note that the SGS without weight decay is the same as OLDA without ``topic mixing", where ``topic mixing" is the key component of OLDA \cite{alsumait2008line}. Moreover, OLDA attempts to solve a different generative model, instead of the usual LDA model. 

The rest of the paper is structured as follows. Section 2 presents the streaming Gibbs sampling (SGS) algorithm for LDA and shows its relationship with Conditional Density Filtering (CDF) \cite{guhaniyogi2014bayesian}.
We also propose the Distributed SGS (DSGS) which can take advantage of sparsity in sampling to handle very large datasets.
Section 3 presents experimental settings of SGS and DSGS to demonstrate their inference quality and speed. Section 4 concludes with discussion on future work. 

\section{Streaming Gibbs Sampling for LDA} 
\subsection{Basics of LDA} 
LDA\cite{blei2003latent} is a hierarchical Bayesian model that describes a generative process of topic proportions for a document and topic assignments for every position in documents, words in the documents are sampled from distributions specified by topic assignments.

Let $D, K, V$ be the number of documents, topic and unique words respectively. $\vec{\phi}_k$ is a $V$ dimensional categorical distribution over words with symmetric Dirichlet prior $\beta$, $\vec{\theta}_d$ is the topic proportion for document $d$ with Dirichlet prior $\alpha$. The generative process of LDA is as follows
\begin{align*}
\vec{\phi}_k \sim Dir(\beta), \forall d \in \{1, \dots, D\}:  \vec{\theta}_d \sim Dir(\alpha), z_{di} \sim Mult(\vec{\theta}_d), w_{di} \sim Mult(\vec{\phi}_{z_{di}}),
\end{align*}
where $L_d$ is the length of document $d$, $i \in \{1, \dots, L_d\}$, $z_{di}$, $w_{di}$ is the assignment and word on position $i$ of document $d$. $Dir(\cdot)$, $Mult(\cdot)$ is the Dirichlet distribution and Multinomial distribution. Denote $\bm{\theta}=(\vec{\theta}_1, \dots, \vec{\theta}_D)$, the matrix formed by all topic proportions, likewise for $\bm{\Phi}, \bm{Z}, \bm{W}$.

Given a set of documents $\bm{W}$, inferring the exact posterior distribution $p(\bm{\theta}, \bm{\Phi}, \bm{Z} | \bm{W})$ is intractable. We must resort to either variational approximation or Monte Carlo simulation methods. Among the various algorithms, collapsed Gibbs sampling (CGS)~\cite{griffiths2004finding} is of particular interest due to its simplicity and sparsity. CGS exploits the conjugacy properties of Dirichlet and Multinomial to integrate out $(\bm{\theta}, \bm{\Phi})$ from the posterior: 
\vspace{-1em}
\begin{equation}
p(z_{di}=k|\bm{Z}^{-di},\bm{W})\propto(N_{kd}^{-di}+\alpha) \frac{N_{kv_{di}}^{-di}+\beta}{N_k^{-di}+V\beta},
\label{eq:CGS_cond}
\end{equation}
where $N_{kd}, N_{kv}$ are sufficient statistics (counts) for the Dirichlet-Multinomial distribution: $N_{kd}=\sum_{i=1}^{L_d} \mathbb{I}(z_{di}=k), N_{kv}=\sum_d\sum_{i=1}^{L_d} \mathbb{I}(w_{di}=v, z_{di}=k), N_k=\sum_d N_{kd}=\sum_v N_{kv}$. The superscript $^{-di}$ stands for excluding the token at position $i$ of document $d$. $\bm{N}_{kv}, \bm{N}_{kd}, \bm{N}_k$ are the matrices or vector formed by all corresponding counts. A pseudocode of CGS is depicted as Alg.~\ref{alg:CGS}. The sparsity of sufficient statistic matrices $\bm{N}_{kw}, \bm{N}_{kd}$ leads to many fast sparsity aware algorithms \cite{yao2009efficient, li2014reducing, yuan2014lightlda}. The sparsity is also an important factor that makes our algorithm more scalable than SVB.
\vspace{-1em}

\begin{figure}[H]
\begin{minipage}[t]{0.5\columnwidth}
\begin{algorithm}[H]
   \caption{Collapsed Gibbs Sampling}
   \label{alg:CGS}
\begin{algorithmic}
   \STATE {\bfseries Input:} data $\bm{W}$, iterations $N$
   \STATE Initialize $\bm{Z}, N_{kv}, N_{k}, N_{dk}$
	
   \FOR{$iter=1$ {\bfseries to} $N$}
      \FOR{each token $z_{di}$ in the documents}
	  			\STATE Sample $z_{di}\sim p(z_{di}|\bm{Z}^{-di},\bm{W})$
  				\STATE Update $N_{kv}, N_k, N_{dk}$
  		\ENDFOR
   \ENDFOR
   \STATE {\bfseries Output posterior mean:} \\
   		$\phi_{kv}=\frac{N_{kv}+\beta}{N_{k}+V\beta}$,
		$\theta_{dk}=\frac{N_{dk}+\alpha}{N_{d}+K\alpha}$
\end{algorithmic}
\end{algorithm}
\end{minipage}
\begin{minipage}[t]{0.5\columnwidth}
\begin{algorithm}[H]
   \caption{Streaming Gibbs Sampling} 
   \label{alg:SGS}
\begin{algorithmic}
   \STATE {\bfseries Input:} iterations $N$, decay factor $\lambda$
   \FOR{$t=1$ {\bfseries to} $\infty$}   		
	   \STATE {\bfseries Input:} data $\bm{W}^t$  
	   \STATE Initialize $\bm{Z}^t$ and update $N_{kv}^t, N_k^t, N_{dk}^t$ 
	   \FOR{$iter=1$ {\bfseries to} $N$}
	        \FOR{each token $z_{di}$ in the mini-batch}
	            \STATE Sample $z_{di}\sim p(z_{di}|\bm{Z}_{-di}^{1:t},\bm{W}^{1:t})$
	            \STATE Update $N_{kv}^{t}, N_k^{t}, N_{dk}^t$
  			 \ENDFOR
	   \ENDFOR
	   \STATE Decay: $\bm{N_{kv}^{t}}=\lambda \bm{N_{kv}^{t}}$
	   \STATE {\bfseries Output posterior mean:} \\
 	   		$\phi_{kv}^t=\frac{N_{kv}^t+\beta}{N_{k}^t+V\beta}$,
			$\theta_{dk}^t=\frac{N_{dk}^t+\alpha}{N_{d}^t+K\alpha}$  		   		
   \ENDFOR   
\end{algorithmic}
\end{algorithm}
\end{minipage}
\end{figure}

\subsection{Streaming Gibbs Sampling}

Given a Bayesian model $P(x|\bm{\Theta})$ with prior $P(\bm{\Theta})$, and incoming data mini-batches $\bm{X}^1, \bm{X}^2, \cdots, \bm{X}^t, \cdots$, let $\bm{X}^{1:t}=\{\bm{X}^1,\dots, \bm{X}^t\}$. 
Bayesian streaming learning is the process of getting a series of posterior distributions $P(\bm{\Theta}|\bm{X}^{1:t})$ by the recurrence relation:
\begin{align}
P(\bm{\Theta}|\bm{X}^{1:t})\propto P(\bm{\Theta}|\bm{X}^{1:t-1})P(\bm{X}^t|\bm{\Theta}).\label{eqn:online}
\end{align}
Therefore, the posterior learnt from $\bm{X}^{1:t-1}$ is used as the prior when learning from $\bm{X}^t$. Note that the amount of data in a stream might be infinite, so a streaming learning algorithm can neither store all previous data, nor update the model at time $t$ with a time complexity even linear of $t$. Ideally, the algorithm should only have constant storage and constant update complexity at each time step.


As depicted in Alg.~\ref{alg:SGS}, we propose a streaming Gibbs sampling (SGS) algorithm for LDA. SGS is an online extension of CGS, which fixes the topics $\bm{Z}^{1:t-1}$ of the previous arrived document mini-batch, and then samples $\bm{Z}^t$ of the current mini-batch using the normal CGS update. This is in contrast with CGS, which can come back and refine some $\bm{Z}^{t}$ after it is first sampled. Actually $\bm{Z}^{1:t-1}$ is not even stored in CGS, we store only the sufficient statistic $\bm{N}_{kv}$. 

One can understand SGS using the recurrence relation (\ref{eqn:online}): without any data, the initial $\vec{\phi}_k$ have parameters $\vec{\beta}_k^0 = \beta$, after incorporating mini-batch $\bm{W}^1$, the parameters is updated to $\vec{\beta}^1_k = \vec{\beta}^0_k + \bm{N_{kv}^1}[k]$, which is used as the prior of consequent mini-batches, where $\bm{N_{kv}^t}[k]$ is the $k$-th row of matrix $\bm{N_{kv}^t}$. In general, SGS updates the prior using the recurrence relation $\vec{\beta}^t_k = \vec{\beta}^{t-1}_k + \bm{N_{kv}^t}[k]$, where $\vec{\beta}^t_k$ is the prior for $\vec{\phi}_k$ at time $t$. 

The decay factor $\lambda$ serves to forget the history. When plugging in the decay factor, the update equation becomes $\vec{\beta}^t_k = \lambda (\vec{\beta}^{t-1}_k+\bm{N_{kv}^t}[k])$. 
$\lambda$ can then be understood as weakening the posterior caused by the previous data. This decay factor would improve the performance of SGS, especially when the topic-word distribution is evolving along the time.

SGS only requires constant memory: it only stores the current mini-batch $\bm{W}^t, \bm{Z}^t$, but not $\bm{W}^{1:t-1}, \bm{Z}^{1:t-1}, \bm{W}^{t+1:\infty}, \bm{Z}^{t+1:\infty}$. The total time complexity is the same as CGS, which is $O(KN|\bm{W}^{1:t}|)$, where $|\bm{W}^{1:t}|$ is number of tokens in mini-batch $1$ to $t$. In practice, we use a smaller number of iterations than that of CGS, because SGS iterates over a smaller number of documents (mini-batch) and thus it converges faster. 

\subsection{Relation to Conditional Density Filtering}


In this section, we consider a special case of SGS, where the decay factor $\lambda$ is set to 1.0. We relate this special case to Conditional Density Filtering (CDF) \cite{guhaniyogi2014bayesian} and show that SGS can be seen as an improved version of CDF framework when applied to the LDA model.

\subsubsection{Conditional Density Filtering}
CDF is an algorithm that can sample from a sequence of gradually evolving distributions. Given a probabilistic model $P(\bm{D^{1:t}}|\bm{\Theta})$, where $\bm{\Theta}=(\theta_1,\cdots, \theta_k)$ is a $k$-dimensional parameter vector and $\bm{D^{1:t}}$ is the data until now, we define the Surrogate Conditional Sufficient Statistics (SCSS) as follows:\
\begin{mydef}
\label{def:SCSS}
[SCSS] Assume $p(\theta_j|\theta_{-j},D_t)$ can be written as $p(\theta_j|\theta_{-j,1},h(D_t, \theta_{-j,2}))$, where $\theta_{-j}=\bm{\Theta} \backslash \theta_j$, $\theta_{-j,1}$ and $\theta_{-j,2}$ are a partition of $\theta_{-j}$ and $h$ is some known function. If $\hat{\theta}_{-j,2}^t$ is a consistent estimator of $\theta_{-j,2}$ at time t, then $C^t=g(C^{t-1}, h(D_t, \hat{\theta}_{-j,2}^t))$ is defined as the SCSS of $\theta_j$ at time $t$ , for some known function $g$. We use $p(\theta_j|\theta_{-j,1}, C^t)$ to approximate $p(\theta_j|\theta_{-j},D^{1:t})$.
\end{mydef}

SCSS is an extension of Sufficient Statistics (SS), in the sense that both of them summarize the historical observations sampled from a class of distributions. SS is accurate and summarizes a whole distribution, but SCSS is approximate and only summarizes conditional distributions. 

If the parameter set of a probabilistic model can be partitioned into two sets $I_1$ and $I_2$, where each parameter's SCSS only depends on the parameters in the other set, then we can use the CDF algorithm (\ref{alg:CDF}) to infer the posterior of the parameters.

\begin{minipage}[t]{0.5\columnwidth} \vskip -0.7cm
    \begin{algorithm}[H]
       \caption{Conditional Density Filtering}
       \label{alg:CDF}
    \begin{algorithmic}
       \FOR{$t=1$ {\bfseries to} $\infty$}
       		\FOR{$s \in \{1,2\}$}
       			\FOR{$j \in I_s$}
       				\STATE $C_{js}^t=g(C_{js}^{t-1}, h(D_t, \hat{\bm{\Theta}}_{-s}))$
       				\STATE Sample $\theta_j \sim p(\theta_j | \theta_{-js}, C_{js}^t)$
       			\ENDFOR
       		\ENDFOR
       \ENDFOR
    \end{algorithmic}
    \end{algorithm}
    \vskip -0.2in
    \begin{algorithm}[H]
       \caption{Distributed SGS (DSGS)}
       \label{alg:DSGS}
    \begin{algorithmic}
       \STATE {\bfseries Input:} iterations $N$, decay factor $\lambda$
       \STATE Initialize $\bm{N}_{kv}=0$
       \FOR{each mini-batch $\bm{W}^t$ at some worker}
       		\STATE Copy global $\bm{N}_{kv}$ to local $\bm{N}_{kv}^{local}$
       		\STATE $\Delta \bm{N}_{kv}^{local}=CGS(\alpha, \beta+\bm{N}_{kv}^{local}, \bm{W}^t)$
       		\STATE Update global $\bm{N}_{kv}=\lambda (\bm{N}_{kv} + \Delta \bm{N}_{kv}^{local})$
       \ENDFOR   
    \end{algorithmic}
    \end{algorithm}
\end{minipage}%
\hskip 0.1in
\begin{minipage}[t]{0.5\columnwidth} \vskip -0.7cm 
\begin{algorithm}[H]
   \caption{CDF-LDA}
   \label{alg:CDF-LDA}
\begin{algorithmic}
   \STATE Initialize: $\bm{\Phi}=\mbox{Uniform}(0, 1), \hat{\bm{N}}_{kv}^0=0$
   \FOR{$t=1$ {\bfseries to} $\infty$}
   		   \STATE {\bfseries Input:} a single document $\vec{w}_{t }$
   		   \STATE Initialize $\vec{z}_{t }$
   		   
   		   \STATE SCSS of $z$: $\hat{\bm{\Phi}}^t=\bm{\Phi}$		   
   		   \FOR{each token $i$ in doc $t$}   
	   		   \STATE Sample $z_{t i} \sim p(z_{ti}|\vec{z}_t^{-ti}, \hat{\bm{\Phi}}^t)$
   		   \ENDFOR
   		   
           \STATE SCSS of $\bm{\Phi}$:  $\hat{N}_{kv}^t=\hat{N}_{kv}^{t-1}+\sum_i I(z_{ti}=k \wedge w_{ti}=v)$
   		   \FOR{$k=1$ {\bfseries to} $K$}
   		   		\STATE Sample ${\vec{\phi}}_k \sim Dir(\hat{\bm{N}}_{kv}^t+\beta)$
   		   \ENDFOR
   \ENDFOR
\end{algorithmic}
\end{algorithm}
\end{minipage}%

Under a semi-collapsed representation of LDA, where $\vec{\theta}_d$ is collapsed, we can partition the parameters into two sets: $I_1=\{\phi_{kv}\}; I_2=\{z_{di}\}$. The conditional distributions are: 
$$p(\bm{\Phi}|\bm{Z}, \bm{W})=\prod_{k=1}^K Dir(\vec{\phi}_k| \bm{N}_{kv}[k]+\beta), \quad p(z_{di}=k| \vec{z}_d^{-di}, \bm{\Phi}, \bm{W}) \propto (N_{kd}^{-di}+\alpha) {\phi}_{kv_{di}}.$$ 

By definition (\ref{def:SCSS}), we can verify that the SCSS of $\bm{\Phi}$ and $\vec{z}_d$ at time $t$ are $\bm{N}_{kv}$ and $\bm{\Phi}$ respectively. Thus we have the CDF solution of LDA as shown in Algorithm (\ref{alg:CDF-LDA}).

\subsubsection{Relationship between SGS and CDF-LDA}
Our SGS method can be viewed as an improved version of CDF-LDA in the following aspects:
\begin{itemize}
\item In CDF-LDA, SCSS $\bm{\Phi}^t$ is directly sampled from a Dirichlet distribution, which unnecessarily introduces extra source of randomness. Replacing the sampling with the expectation $\hat{\phi}_{kv}^t=\frac{\hat{N}_{kv}^{t-1}+\beta} {\hat{N}_k^{t-1}+V\beta}$ gives better performance in practice. This corresponds to a fully collapsed representation of LDA. It's more statistical efficient due to the Rao-Blackwell Theorem. 
\item The CDF-LDA's sampling update of $z_{ti}$ does not include other tokens in the current document $t$, which could be improved by taking the current document into account. This is especially useful for the beginning iterations, because it enables the topics' preference of doc $t$ to be propagated immediately. 
\item It is hard to say how a single document can be decomposed into topics without looking at the other documents, but this is the case that occurrs at the beginning of CDF-LDA. This would result in inaccurate $z_{ti}$ assignments and therefore polluting $\bm{N}_{kv}$, and finally resulting in low convergence rate on the whole. Our method avoids this problem by processing a mini-batch of documents at a time and allows for multiple iterations over the mini-batch. This would enable topic-assignments to be propagated locally. 
\end{itemize}

To sum up, SGS without weight decay can be seen as an improvement over CDF-LDA not only by adapting a fully collapsed representation, but also by enabling a timely and repeatedly cross-document information flow.


\subsection{Distributed SGS}
Many distributed CGS samplers exist for LDA, including divide-and-conquer~\cite{newman2007distributed}, parameter server~\cite{ahmed2012scalable, li2014scaling} and model parallel~\cite{guhaniyogi2014bayesian} approaches. Although some of them do not have theoretical guarantees, they all perform pretty well in practice. 
Here, we adopt the parameter server approach and present a distributed SGS sampler. 

Same as CGS, the global parameter in SGS is the topic word count matrix $\bm{N}_{kv}$. SGS can be viewed as a sequence of calls to the CGS procedure
$$\Delta \bm{N}_{kv}^t=CGS(\alpha, \beta +\bm{N}_{kv}^{t-1}, \bm{W}^t),$$
where the $k$th row of $\beta + \bm{N}^{t-1}_{kv}$ is the prior of $\vec{\phi}_k$. Then the topic word count matrix can be updated by $\bm{N}_{kv}^t=\lambda(\bm{N}_{kv}^{t-1}+\Delta \bm{N}_{kv}^t)$.

In the parameter server architecture, we store $\bm{N}_{kv}$ in a central server. Each worker fetches the most recent parameter $\bm{N}_{kv}$, runs $CGS(\alpha, \beta +\bm{N}_{kv}^{t-1}, \bm{W}^t)$ to get the updates $\Delta \bm{N}_{kv}^t$, and pushes back the update to the server. Upon receiving an update, the server updates its parameters. In our implementation, the workers run in a fully (hogwild) asynchronous fashion. Although workers may have slightly stale parameters compared to the serial version, affecting the convergence, this hogwild approach is shown to work well in~\cite{newman2007distributed,ahmed2012scalable} as well as our experiments, based on the fact that $\bm{N}_{kv}$ changes slowly. Better schemes such as stale synchronous parallel~\cite{ho2013more} might be used, but it is just a matter of choosing parameter server implementations. A pseudo-code of is given as Alg.~\ref{alg:DSGS}.

In experiments, we can see that this empirical parallel framework can almost linearly scale up SGS with neglectable precision loss. 
Due to the sparseness of $\bm{N}_{kv}$, Distributed SGS (DSGS) has much less communication overhead between the master and workers, hence more scalable than SVB.


\section{Experiments}

We evaluate the inference quality and computational efficiency of SGS. We also assess how the parameters such as mini-batch size, decay factor and the number of iterations affect the performance. We compare with the online variational Bayes approach SVB~\cite{broderick2013streaming}, which has proven to have high inference quality that is similar to offline  stochastic methods like SVI~\cite{hoffman2013stochastic}. 

\subsection{Implementation Details}\label{sec:initialization}

As Canini et al.~\cite{canini2009online} mentioned, initializations might have big impacts on the solution qualities of the inference algorithms, and hence, using randomized initialization for $\bm{Z}$ is often not good. Thus, we use a kind of ``progressive online" initialization for SGS, CGS and SVB. Specifically, taking SGS as an example, at the first iteration for each mini-batch, for each document we sample $\vec{z_d}$ from the posterior distribution up to current $t$. Then we update the posterior distribution using the current document and proceed to the next one. Such a ``sampling from posterior" initialization technique ensures that our initialization is reasonably good. We use a similar initialization method for SVB and CGS. 


All the core implementations (sampling and calculating a variational approximation) are in C++. We also use Python and MATLAB wrappers for computationally inexpensive operations, such as measuring time. Our implementation of SVB has been made as similar as possible with SGS for fair speed comparison. The speed of our SVB implementation is similar to the implementation in \cite{broderick2013streaming}. The experiments are done on a 3 node cluster, with each node equipped with 12 cores of Intel Xeon E5645@2.4GHz, 24GB memory and 1Gbps ethernet. 

In the distributed experiments, data $\bm{W}^t$ is pre-partitioned and stored separately on each node, but in practice it can be stored in a distributed file system such as HDFS, or a publish-subscribe pattern can be used for handling streaming data. For the sake of simplicity, we implement our own parameter server using pyRpc, a simple remote procedure call (RPC) module that uses ZeroMQ. 
We use a pipeline on worker nodes to hide the communication overhead. Parameters on master server is stored using \texttt{atomic}, and there are model replicas on master server to ensure high availability while performing updates. 
For production usage, existing high performance parameter servers, such as \cite{li2014scaling} might be used to achieve better performance and scalability. Again, system is orthogonal with the algorithm in our case and is not the main focus in this paper.

\subsection{Setups} \label{sec:exp_setting}

In the following experiments, hyper-parameters $\alpha$ and $\beta$ are all set to $0.1$ and $0.03$, number of topics $K$ is set to $50$. Different settings yield same conclusions. Thus we stick to this setting for simplicity. Multiple random starts of SGS and SVB don't result in significant difference. Without special remarks, for each mini-batch, both SGS and SVB run the sampler until convergence. To be specific, SGS stops the iteration when the training perplexity on the current mini-batch stops improving for 10 consecutive iterations, or when it reaches a maximum of 400 iterations. SVB stops the inner iteration when $\frac{||\vec{\theta}_d^{old} - \vec{\theta}_d^{new}||_1}{K} < 10^{-5}$ or when it reaches a maximum of 100 iterations, and stops the outer iteration if
$\frac{||\bm{\phi}^{new}-\bm{\phi}^{old}||_1}{KV} < 10^{-3}$.

The predictive performance of LDA can be measured by the probability it assigns to the held-out documents. This metric is formalized by perplexity. First we partition the data-set and use 80\% for training and 20\% for testing. Let $\mathcal{M}$ denote the model trained on the training data. Given a held-out document $\vec{w}_{d}$, we can infer $\vec{\theta}_d$ from the first half of the tokens in $d$, and then calculate the exponentiated negative average of the log probability. Formally, we have:
$$per(\vec{w}_{d}|\mathcal{M})=\exp\left\{ -\frac{\sum_i \log p(w_{di}|\mathcal{M})}{|\vec{w}_{d}|} \right\}, $$
where $\log p(w_{di}|\mathcal{M})=\log {\sum_k \phi_{kv_{di}}\theta_{dk}}$.

\begin{wraptable}{r}{0.5\textwidth}\vspace{-.5cm}
\vspace{-0.5cm}
\caption{Data Statistics, where $K$ and $M$ stand for thousand and million respectively. }
\label{tb:datasets}
\begin{tabular}{l | lll}
\hline
       & \# Docs  & \# Token & Vocab-Size \\ 
       \hline
NIPS   & 1740 & 2M    & 13K   \\
NYT    & 300K & 100M  & 100K  \\
PubMed & 8.2M & 730M  & 141K  \\
\hline 
\end{tabular}
\vskip -0.5cm
\end{wraptable}

To compare the performance of the algorithms in various settings, we test them on three datasets \footnote{All the three datasets can be downloaded from UCI Machine Learning Repository: \url{https://archive.ics.uci.edu/ml/datasets/Bag+of+Words}}. The small NIPS dataset has the articles from 1988 to 2000 published by Neural Information Processing Systems (NIPS) Conferences. A larger dataset is the news from New York Times (NYT) and the largest one is the abstracts of the publications on PubMed~\cite{UCI+Bache+Lichman:2013}. Table~\ref{tb:datasets} shows the detailed information of each dataset.

\subsection{Results for various mini-batch sizes}

\begin{figure}[t]\vspace{-.4cm}
\centering 
    \subfigure[Perplexities of SVB and SGS on NIPS.] { 
    \label{fig:batch_nips} 
    \includegraphics[width=0.45\columnwidth, height=4cm]{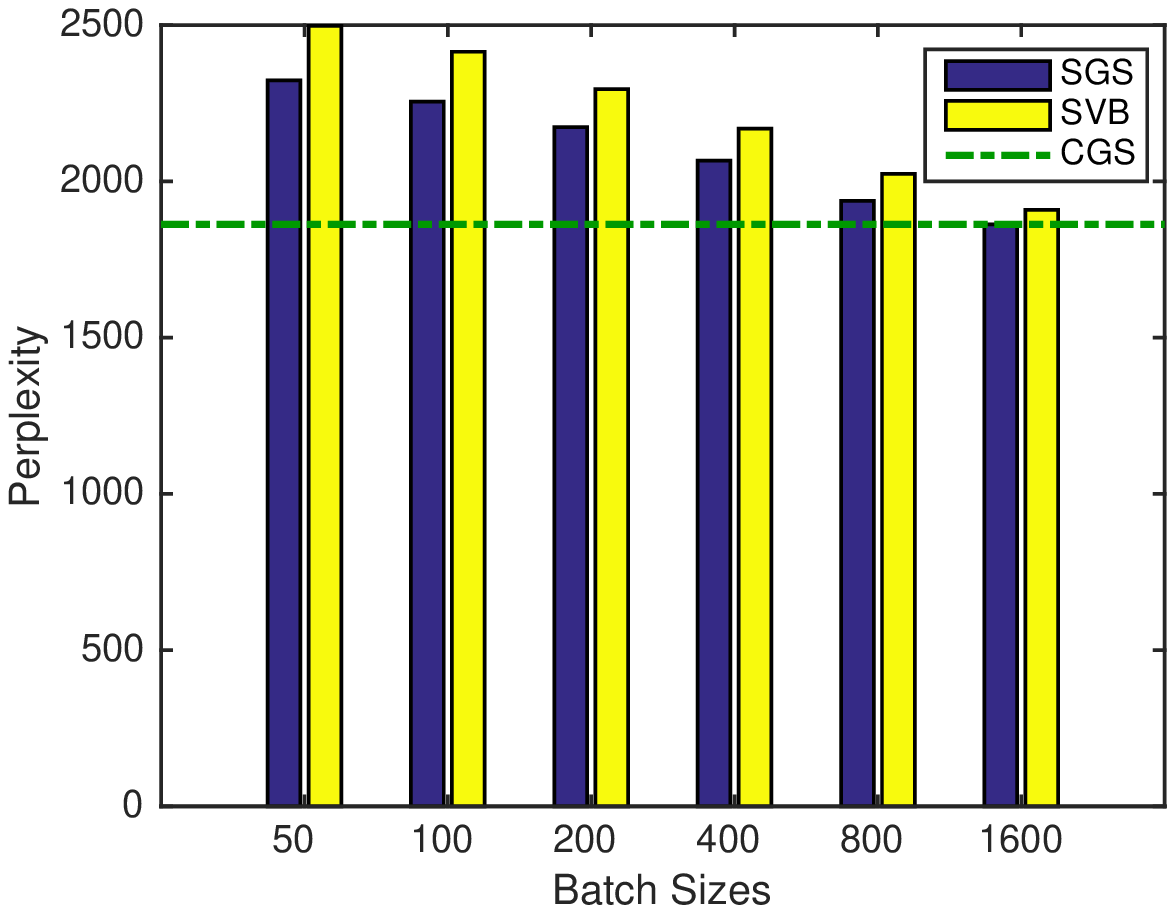}
    }
    \subfigure[Perplexities of SVB and SGS on NYT.] { 
    \label{fig:batch_nyt} 
    \includegraphics[width=0.45\columnwidth, height=4cm]{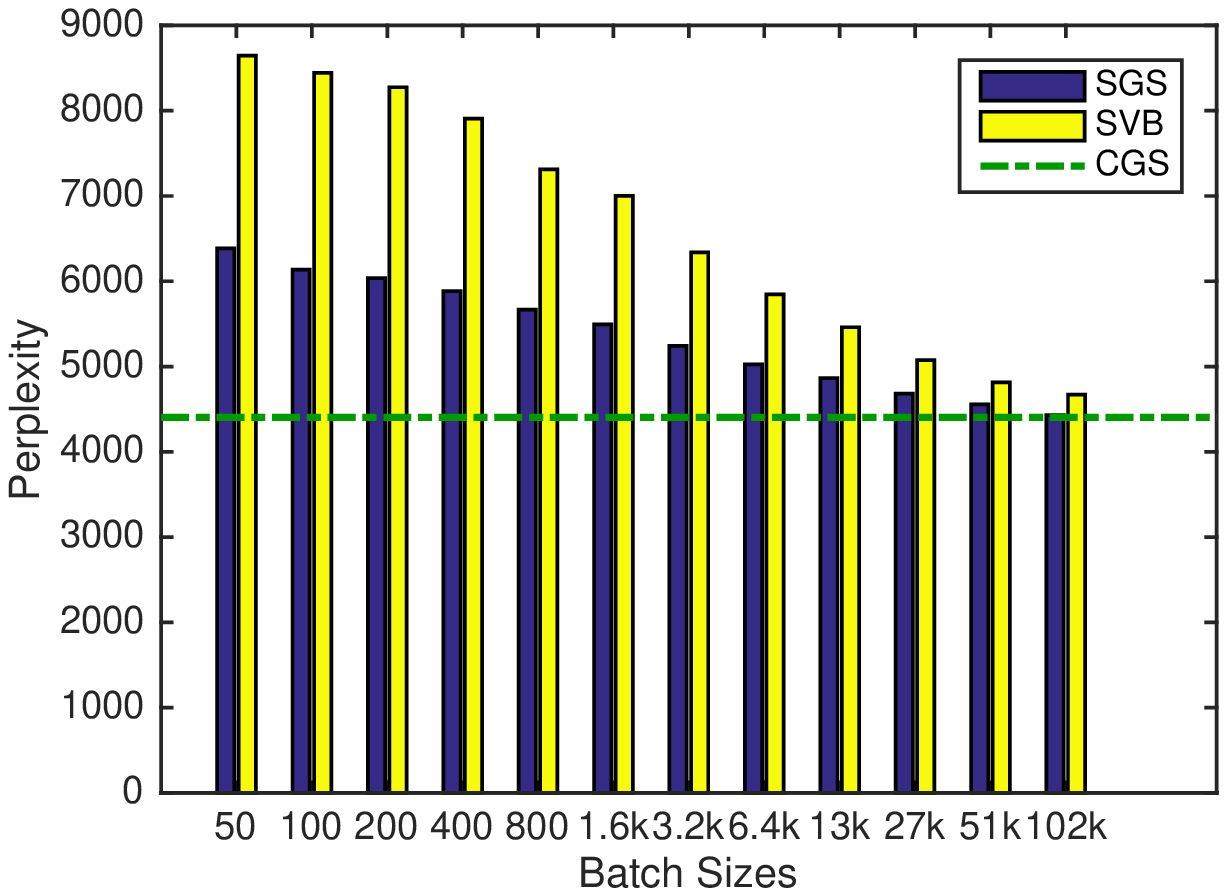} 
    }\vspace{-.3cm}
\caption{The perplexities of SVB, SGS and CGS on the small NIPS (a) and large NYT (b) datasets. }
\label{fig:batch_small} \vspace{-.4cm}
\end{figure}
We run SGS and SVB on NIPS and NYT datasets, varying mini-batch sizes. To simplify our comparison, we set the decay factor for SGS to $\lambda=1$. Fig.  \ref{fig:batch_small}(a) and Fig.  \ref{fig:batch_small}(b) show
that SGS consistently performs better than SVB, especially in the cases of bigger dataset and smaller mini-batch sizes. 

The difference in the performance gap between SGS and SVB on NYT and NIPS datasets could be understood as different levels of redundancy. In order learn effectively in a streaming setting, it is required to have a redundant dataset. There are only 1740 documents in the NIPS corpus, which is far from being redundant and thus different streaming algorithms performs alike on this dataset. From the trend of the dataset size, we can expect that SGS would perform even better than SVB on larger datasets. Another interesting phenomenon is that SVB is more sensitive to mini-batch sizes than SGS. The inherent ability of SGS to perform much better than SVB on smaller mini-batches have important advantages in practice.


Note that SGS is equivalent to CGS when the mini-batch size equals to the whole training set size. The green horizontal dashed lines in Fig. \ref{fig:batch_nips} \ref{fig:batch_nyt} mark the perplexity of CGS. We can see that on the large NYT dataset, SGS has a huge improvement over SVB when compared to the best-achievable perplexity. 



\subsection{Results for different decay factor}
\begin{figure}[t]\vspace{-.3cm}
\begin{center}
    \subfigure[Effect of decay factors (NIPS).] { 
    \label{fig:decay_nips} 
    \includegraphics[width=0.3\columnwidth] {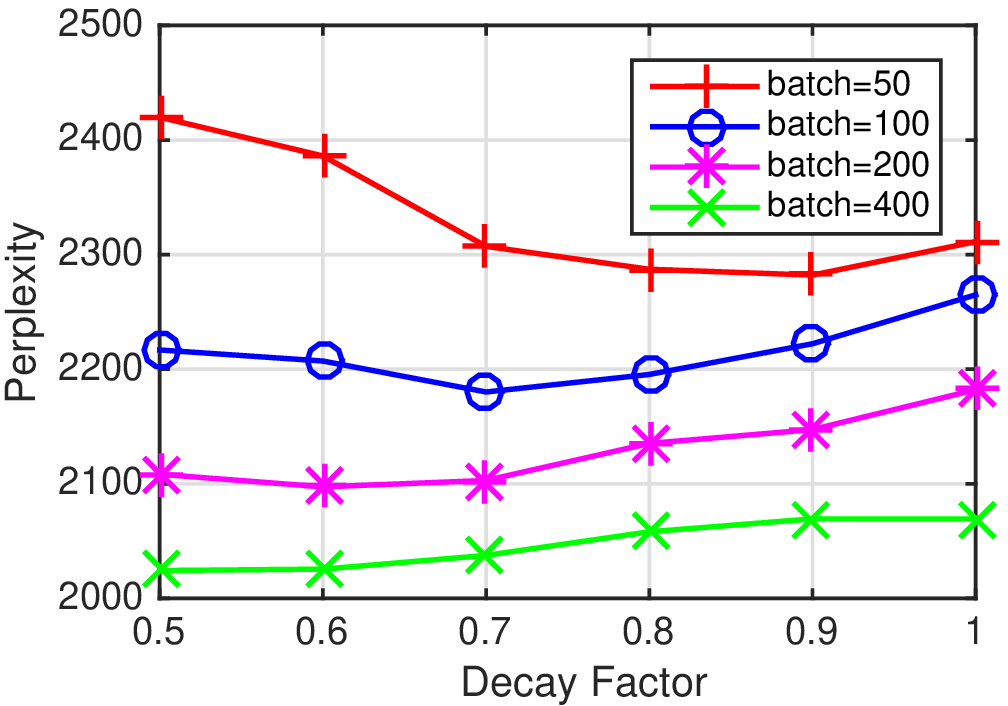}
    }
    \hspace{-0.1in}
    \subfigure[Effect of decay factors (NYT).] { 
    \label{fig:decay_nyt} 
    \includegraphics[width=0.3\columnwidth] {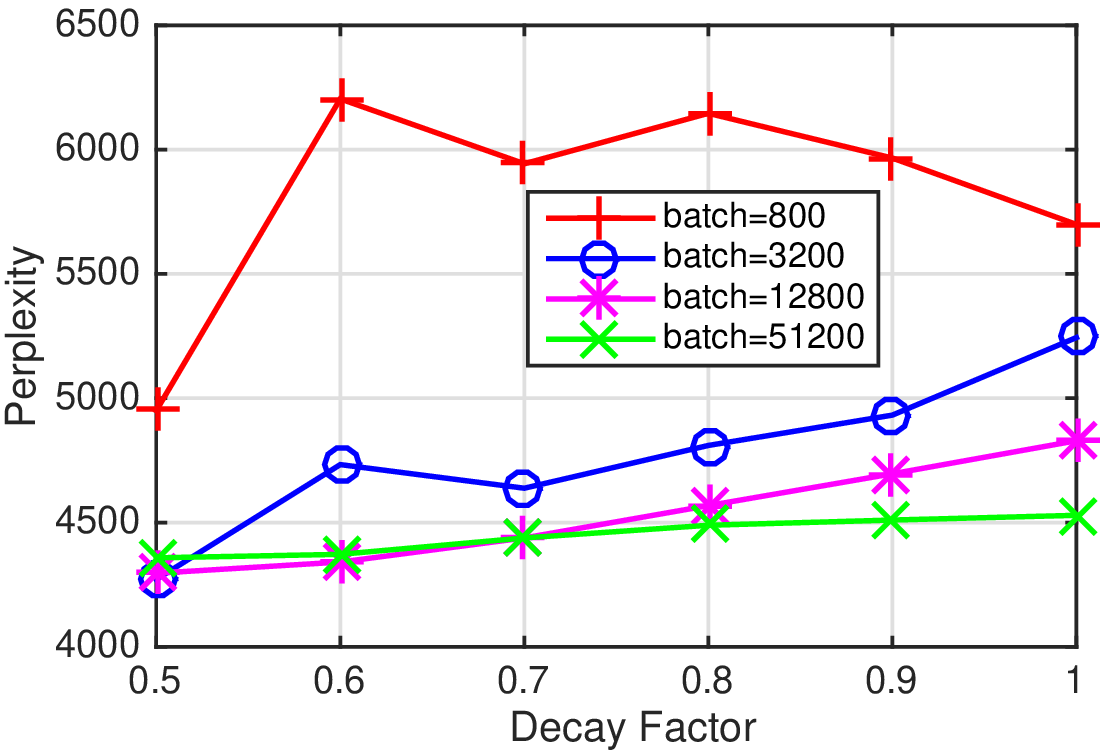}
    }
    \hspace{-0.1in}
    \subfigure[Learning process.] { 
    \label{fig:pro} 
    \includegraphics[width=0.3\columnwidth] {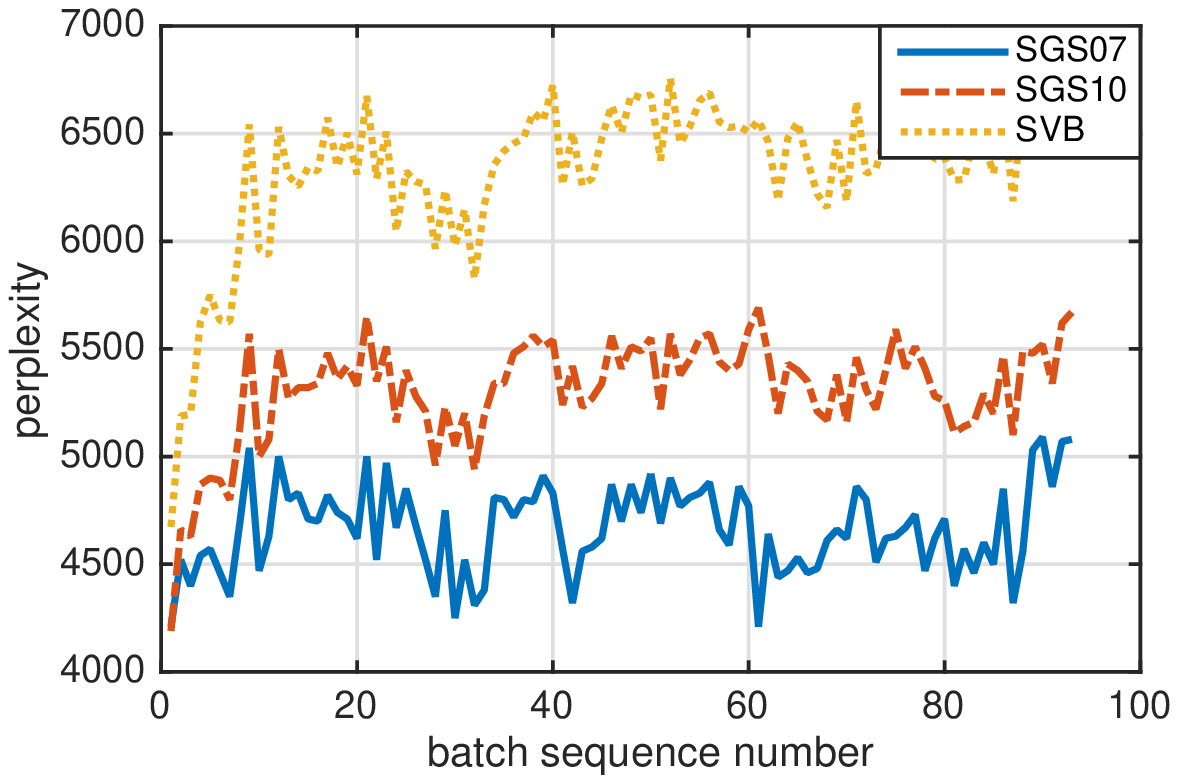}
    }\vspace{-.3cm}
\caption{(a,b) The change of SGS's perplexity w.r.t the decay factor $\lambda$ on NIPS and NYT datasets; and (c) The trend of changing perplexities as new mini-batches arrive, mini-batch size = 3200. }
\end{center}\vspace{-0.5cm}
\end{figure}

We also investigate how the decay factor $\lambda$ affects the performance of SGS on different datasets. The results are similar as above, a bigger dataset has more obvious trends. As shown in Fig.s~\ref{fig:decay_nips} and \ref{fig:decay_nyt}, when the mini-batch is too small, the decay factor has a negative effect. 
It is probably because a small mini-batch can only learn a limited amount of knowledge in a single round and it is thus not preferable to forget the knowledge.
When the mini-batch size gets bigger, the decay factor improves the performance, and the optimal value for $\lambda$ gets smaller. This can be explained in a similar manner as  above, where bigger mini-batches will learn more and the next mini-batch would have greater discrepancy with the current one. 

Let us examine a specific setting where the batch size of NYT dataset is set to $3200$. SVB yields a perplexity $6511$, while SGS without decay can reach $5240$. After applying decay factor of $0.7$, SGS yields a perplexity of $4640$, which is pretty close to the batch perplexity of $4300$. We can conclude that, if the decay factor is set properly, SGS is much better than SVB and it can almost reach the same precision as its batch counterpart.


\subsection{Learning Process}
However, the mean perplexity of each mini-batch is not a full description of the learning process. We should also take the trend of the inference quality as new mini-batches arrive into account. In Fig.~\ref{fig:pro}, we partition each mini-batch into training and testing tests, and plot the testing perplexity of each mini-batch of SVB and SGS. We can clearly see that the performance of the decayed SGS is strictly better than the non-decayed version, and the latter one outperforms SVB. In other words, SGS can consistently learn a better model every day. All three models have perplexity bursts at a few initial mini-batches because the models over-fits the first few mini-batches. 

\subsection{Computational Efficiency}
\begin{figure}[ht] \vskip -0.2cm
\begin{center}
    \subfigure[Convergence of SGS and SVB on a mini-batch.] { 
    \label{fig:test_conv} 
    \includegraphics[width=0.45\columnwidth]{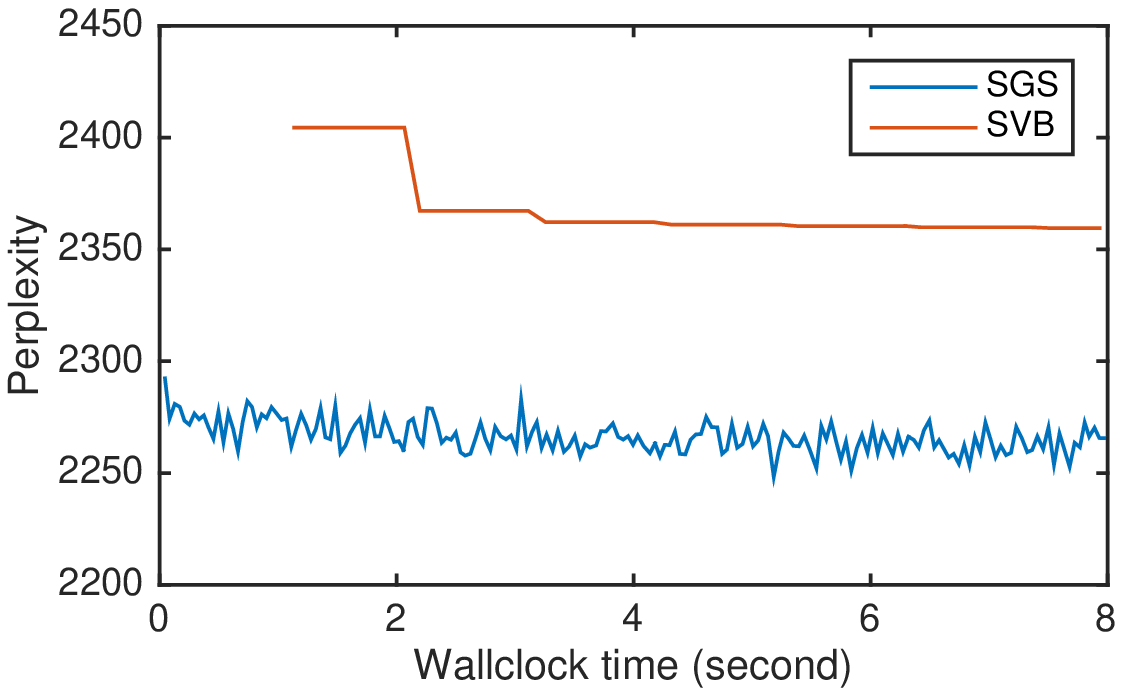}
    }
    \hspace{-0.1in}
    \subfigure[Time consumption of SGS, SVB and CGS.] { 
    \label{fig:time_effi} 
    \includegraphics[width=0.45\columnwidth]{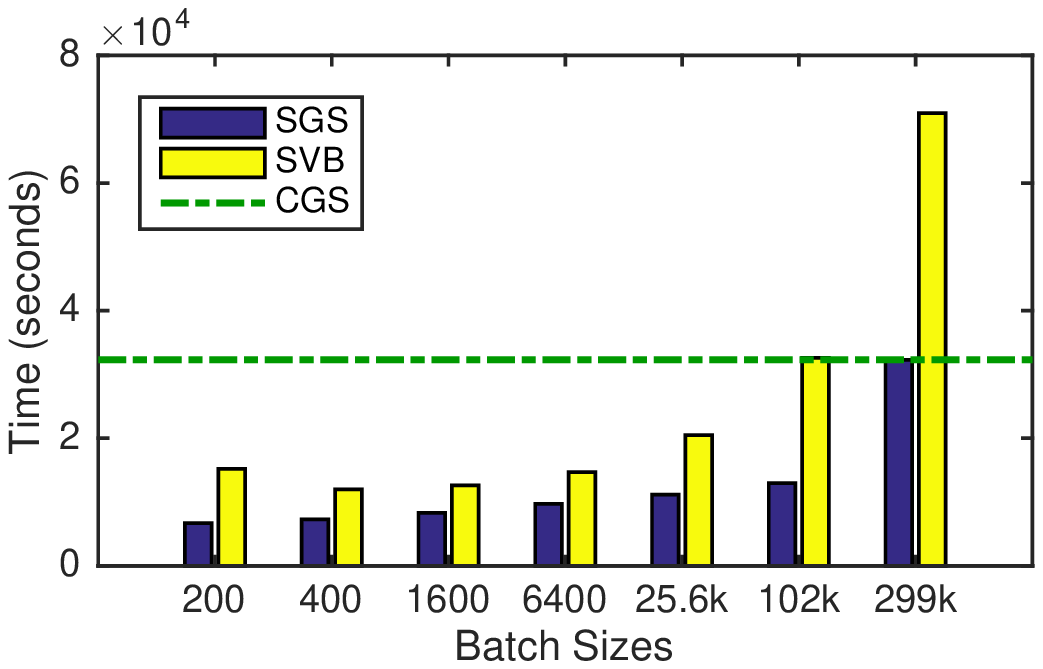}
    }
\vskip -0.2cm
\caption{
Computational Efficiency of SGS, SVB, and CGS
}
\end{center}
\vskip -0.4cm
\end{figure}

Since online algorithms usually run for a longer time span, months or years, online algorithms face less computational challenges than the offline versions. However, we would still like the online inference method to not use excessive computational resources. In this section, we compare the computational efficiencies of SGS, SVB and CGS. Since SGS and SVB have outer loops that process arriving mini-batches and inner loops (iterations), we investigate the time per mini-batch and on the whole dataset separately.

In Fig.~\ref{fig:test_conv}, we run SGS and SVB through some initial mini-batches and then investigate their convergences on the same intermediate mini-batch. The perplexity on held-out documents are plotted against time. SGS starts from a better point than SVB because its better result on previous iterations. Furthermore, SGS also converges faster than SVB. 

In Fig.~\ref{fig:time_effi}, both SGS and SVB are run until convergence, where the criterion for convergence is stated in Section \ref{sec:exp_setting}. We can see that SVB and SGS converge within similar time. Also, since the online version is searching over smaller number of configurations, we can observe that the smaller the mini-batch size is, the faster it converges. SGS can be faster than CGS with smaller mini-batch sizes.

\subsection{Distributed Experiments}
\begin{figure}[t] \vskip -0.5cm
\centering 
    \subfigure[Perplexity of DSGS and SVB on NYT.]{
    \label{fig:nyt_pers} 
    \includegraphics[width=0.45\columnwidth, height=3.5cm]{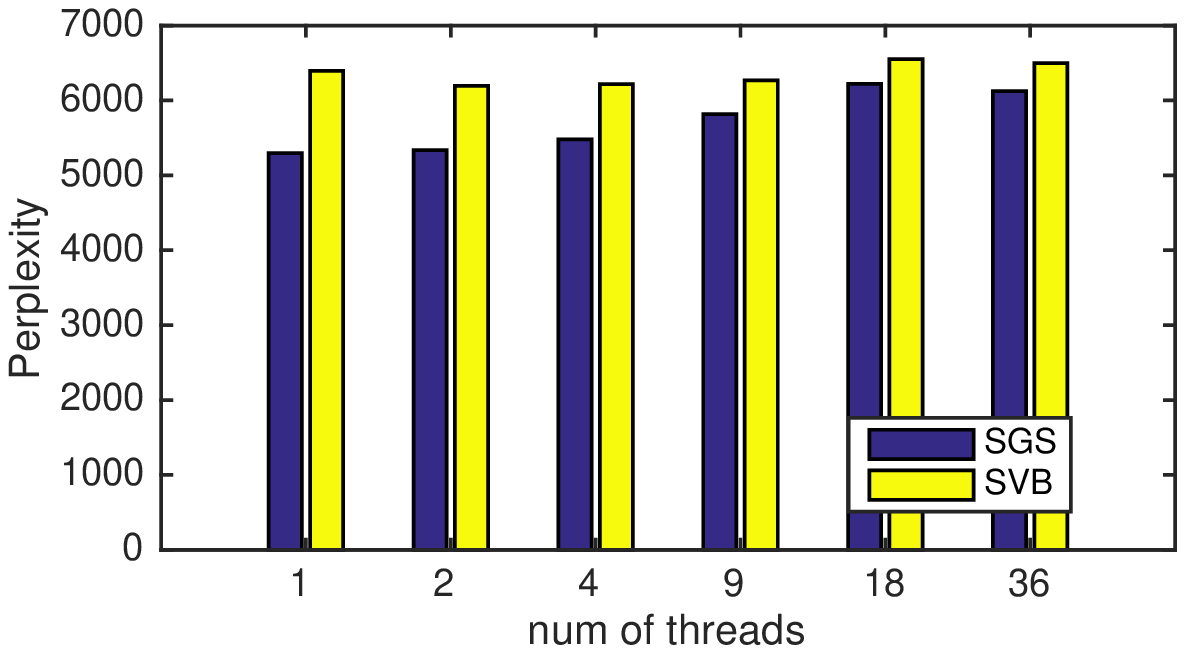}
    }
    \subfigure[Tokens per second of DSGS and SVB.]{
    \label{fig:nyt_time} 
    \includegraphics[width=0.45\columnwidth, height=3.5cm]{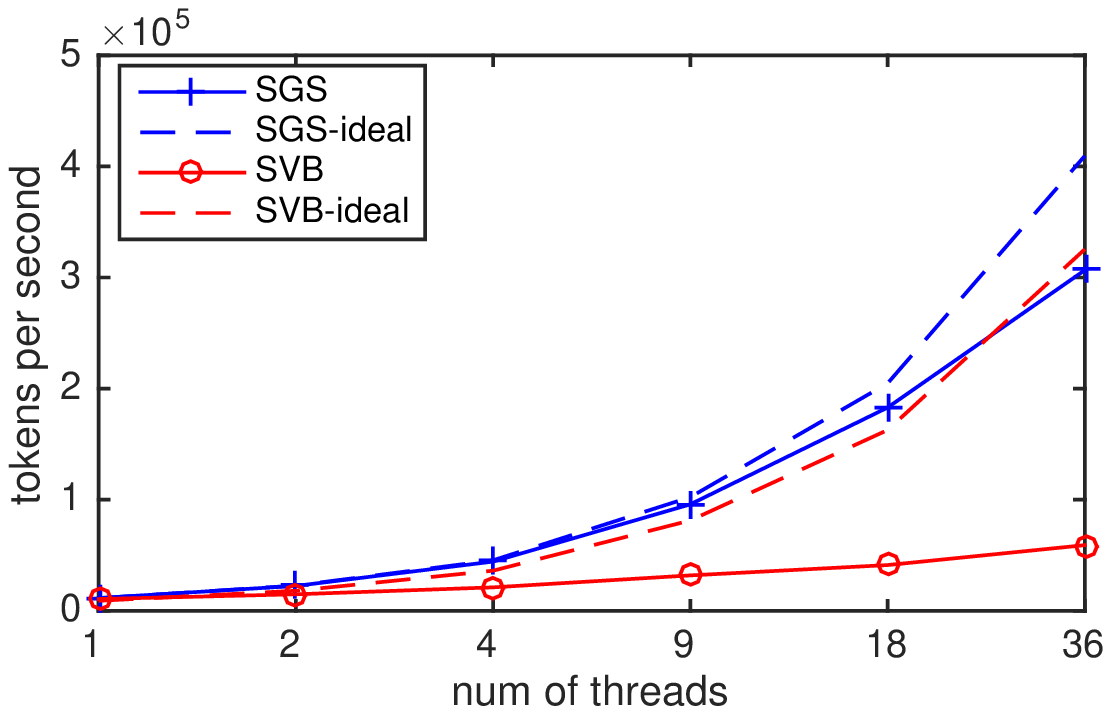} 
    }
    
    \subfigure[Perplexity of DSGS on PubMed dataset.] { 
    \label{fig:dis_precision} 
    \includegraphics[width=0.45\columnwidth, height=4cm]{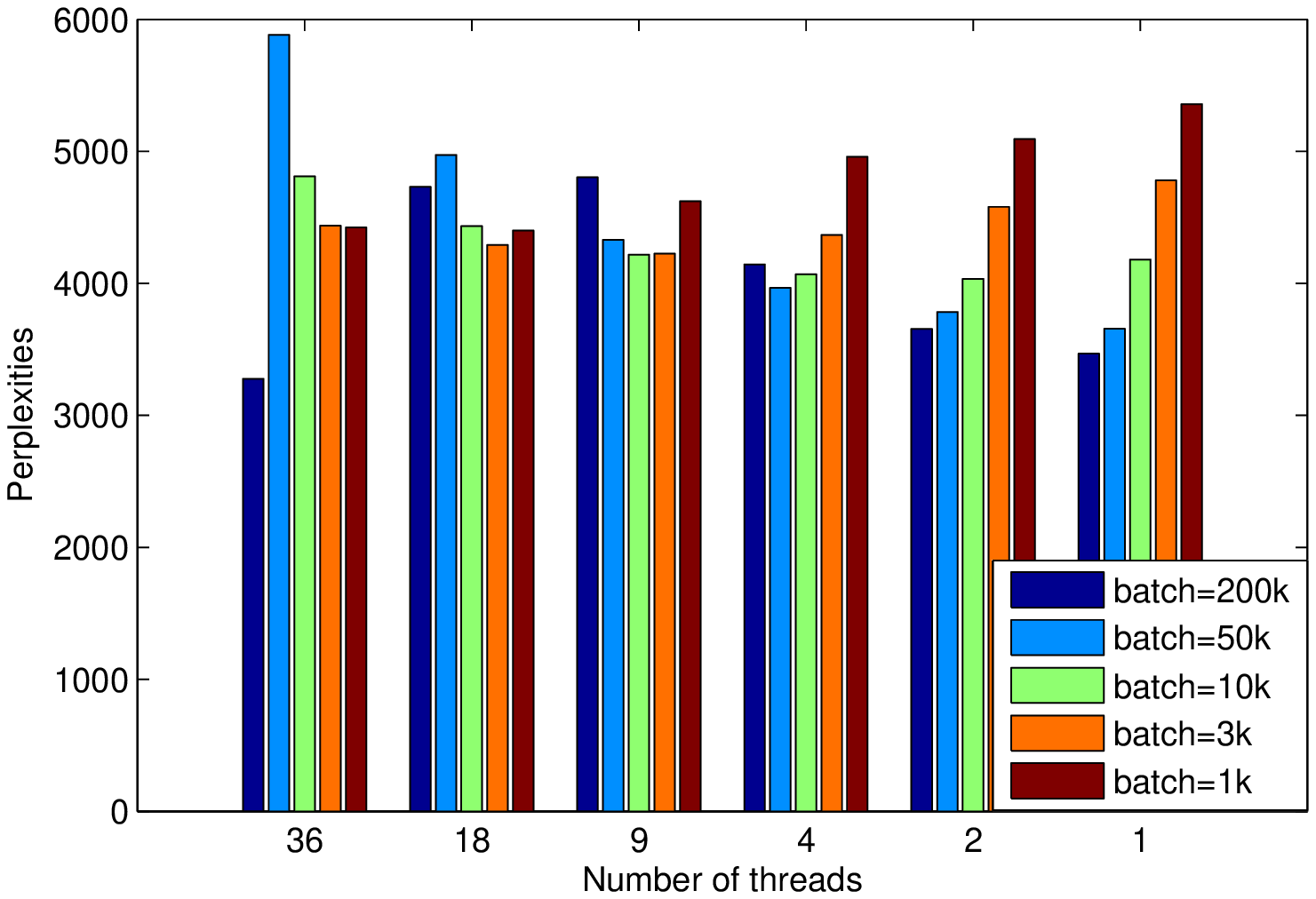} 
    }
    \subfigure[Time consumption of DSGS on PubMed.] { 
    \label{fig:dis_time} 
    \includegraphics[width=0.45\columnwidth, height=4cm]{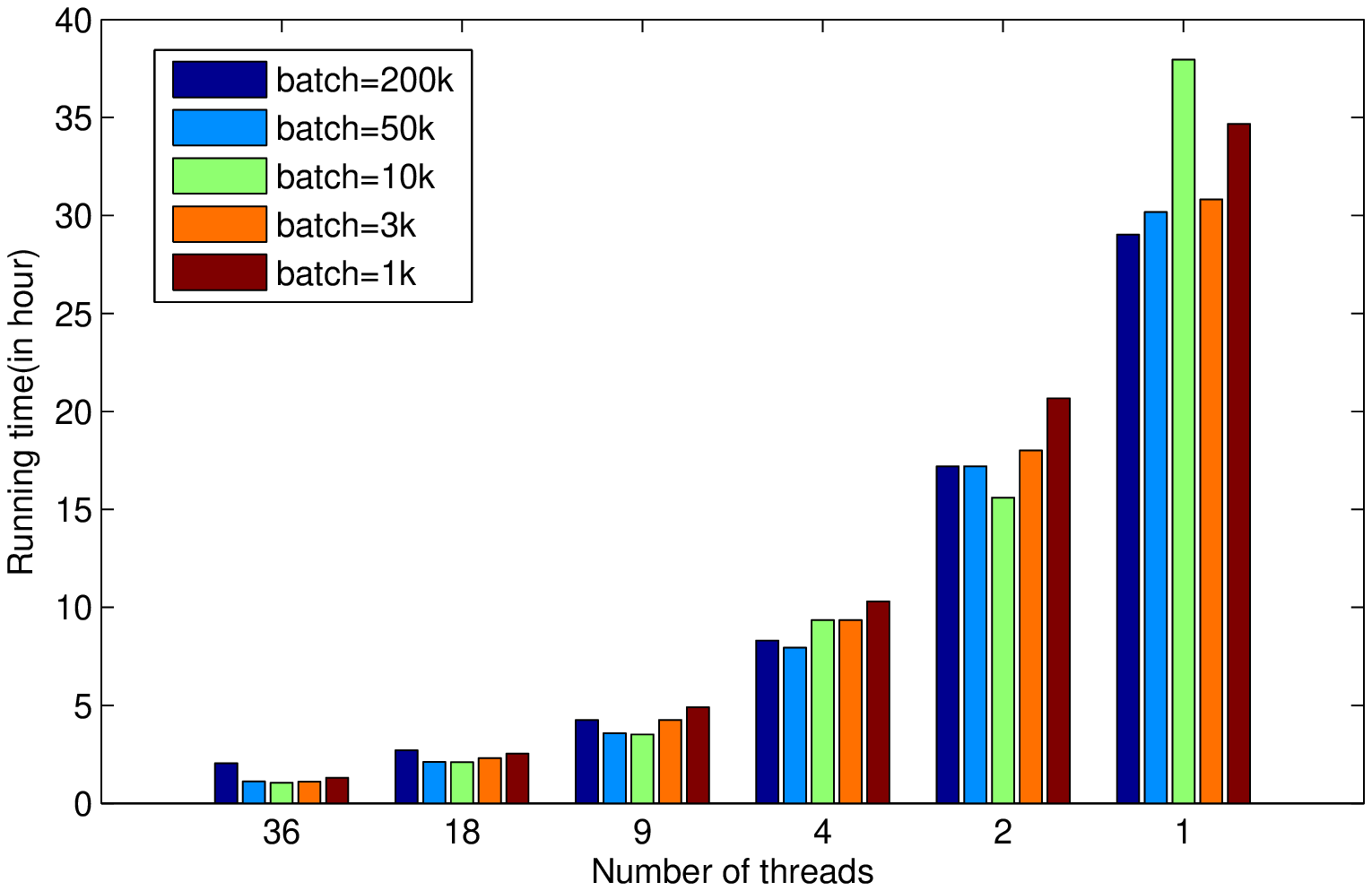} 
    }
\vskip -0.2cm
\caption{Scalability results }
\label{fig:batch} 
\vskip -0.5cm
\end{figure}

In this section we compare distributed SGS and SVB\footnote{SVB refers to both distributed and single-threaded variants.} on the NYT dataset, and the scalability of DSGS is examined on the larger PubMed dataset. 
When we compare DSGS with SVB in Fig.~\ref{fig:nyt_pers}, we can conclude that although the perplexity of DSGS gets worse as the number of cores increases, it still consistently outperforms SVB. Fig.~\ref{fig:nyt_time} shows the throughput of DSGS and SVB, in tokens per second.
Since the topic-word assignment update of DSGS is sparse and the corresponding variational parameter update of SVB is dense, the speedup of DSGS is much better than SVB. Fig.~\ref{fig:dis_precision},~\ref{fig:dis_time} show the scalability result on the larger PubMed dataset. In general we can conclude DSGS enjoys nice speedup while retaining a similar level of perplexity.

\section{Discussion}
We have developed a streaming Gibbs sampling algorithm (SGS) for the LDA model. Our method can be seen as an online extension of the collapsed Gibbs sampling approach. 
Our experimental results demonstrate that SGS improves perplexity over previous online methods, while maintaining similar computational burden. We have also shown that SGS can be well parallelized using similar techniques as those adopted in SVB. 

In the future, SGS can be further improved by making the decay factor $\lambda$, the mini-batch size and the number of iterations for each document self-evolving, as more data is fed into the system. Intuitively, the algorithm learns fast at the beginning and slows down later on. Thus for example, it might be tempting to decrease the iteration counts for each document to some constant over time. The scheme for the evolution deserves future research and needs strong theoretical guidance. 


{\small
\printbibliography
}

\end{document}